\pdfoutput=1

\documentclass[11pt]{article}

\usepackage[table]{xcolor}
\usepackage[preprint]{acl}

\usepackage{times}
\usepackage{latexsym}

\usepackage[T1]{fontenc}

\usepackage[utf8]{inputenc}

\usepackage{microtype}

\usepackage{inconsolata}

\usepackage{graphicx}

\usepackage{hyperref}       
\usepackage{url}            
\usepackage{booktabs}       
\usepackage{tabularx}
\usepackage{multirow}
\usepackage{amsfonts, amsmath, amssymb}       
\usepackage{nicefrac}       
\usepackage{xcolor}         

\usepackage{wrapfig}
\usepackage{subcaption}
\usepackage{array}
    \newcolumntype{P}[1]{>{\centering\arraybackslash}p{#1}}
    \newcolumntype{M}[1]{>{\centering\arraybackslash}m{#1}}
\usepackage{makecell}

\DeclareMathOperator*{\argmax}{\arg\!\max}

\usepackage{amsthm}

\usepackage[capitalize]{cleveref}
\crefname{table}{Table}{Tabs.}
\crefname{figure}{Fig.}{Figs.}
\Crefname{section}{Section}{Sections}
\Crefname{table}{Table}{Tables}
\Crefname{assumption}{Assumption}{Assumptions}
\crefname{algorithm}{Algorithm}{Algorithms}

\author{Karuna Bhaila \quad Minh-Hao Van \quad Xintao Wu \\
        University of Arkansas \\
        \texttt{\{kbhaila, haovan, xintaowu\}@uark.edu}}

\title{Soft Prompting for Unlearning in Large Language Models}

\begin{document}
\maketitle
\begin{abstract}
The widespread popularity of Large Language Models (LLMs), partly due to their unique ability to perform in-context learning, has also brought to light the importance of ethical and safety considerations when deploying these pre-trained models. In this work, we focus on investigating machine unlearning for LLMs motivated by data protection regulations. In contrast to the growing literature on fine-tuning methods to achieve unlearning, we focus on a comparatively lightweight alternative called soft prompting to realize the unlearning of a subset of training data. With losses designed to enforce forgetting as well as utility preservation, our framework \textbf{S}oft \textbf{P}rompting for \textbf{U}n\textbf{l}earning (SPUL) learns prompt tokens that can be appended to an arbitrary query to induce unlearning of specific examples at inference time without updating LLM parameters. We conduct a rigorous evaluation of the proposed method and our results indicate that SPUL can significantly improve the trade-off between utility and forgetting in the context of text classification and question answering with LLMs. We further validate our method using multiple LLMs to highlight the scalability of our framework and provide detailed insights into the choice of hyperparameters and the influence of the size of unlearning data. 
Our implementation is available at \url{https://github.com/karuna-bhaila/llm_unlearning}.  
\end{abstract}

\section{Introduction}
With advancements in transformer models~\citep{vaswani2017attention} and the availability of massive text corpus, language models have rapidly evolved over the past decade. The \textit{pre-train and fine-tune} pipeline has garnered wide popularity, especially since the release of LLMs such as GPT~\citep{openai2024gpt4} and LLaMA~\citep{touvron2023LLaMA}.
However, ethical and security concerns have been raised due to the inclusion of private and sensitive information in the training data. For example, LLMs can regurgitate individual personal information~\citep{nasr2023scalable}, or mimic harmful and/or hateful behavior as a consequence of such content being prevalent in the data~\citep{wen2023toxicity}. The non-consented and unwarranted use of copyrighted content for LLM training has also raised significant concerns~\citep{eldan2023harry, nyt2023sue}.

Current policies governing the use and distribution of such models do not encompass all ethical avenues; nonetheless, certain regulations such as California Consumer Privacy Act (CCPA) and GDPR's Right to be Forgotten (RTBF) serve as guidelines for organizations to ensure that their operations do not infringe upon user privacy. Specifically, these regulations stipulate that businesses and data collectors provide and exercise an \textit{opt-out} mechanism essentially allowing individuals to request the deletion of their data on reasonable grounds.
In machine learning literature, these regulations have been conceptualized as machine unlearning~\citep{cao2015unlearning, bourtoule2021unlearning}, which aims to eliminate the influence of unwanted data points on a model's behavior as if they had never been observed during training.  
Naturally, machine unlearning should be integrated into the LLM pipeline to address the previously outlined issues resulting from the presence of sensitive data in pre-training. However, unlearning in LLMs faces unique challenges due to the inaccessibility of model and pre-training data, and the sheer size of the pre-trained LLMs making re-training practically infeasible. Much of the research in this direction therefore focuses on the fine-tuning approach which involves training all or a subset of LLM parameters to enforce unlearning~\citep{jang2023unlearning, chen2023efficientunlearning, yao2024large, maini2024tofu, yao2024machine}.  

In this work, we propose a novel approach to unlearning in LLMs via soft prompting which is less resource-intensive than fine-tuning. To the best of our knowledge, this is the first work to investigate the use of soft prompting for unlearning in LLMs. Soft prompting simplifies the process of adapting LLMs to an arbitrary downstream task by optimizing learnable token embeddings that encode signals from a corresponding dataset~\citep{lester2021prompttuning, li2021ptuning}. The soft prompts are trained end-to-end and essentially act as instructions for a frozen pre-trained LLM during inference. We leverage this ability to modulate LLM outputs using prompts and formulate \textbf{S}oft \textbf{P}rompting for \textbf{U}n\textbf{l}earning (SPUL), a resource-efficient mechanism to achieve LLM unlearning in text classification and multiple-choice question answering (MCQA). 
We optimize a set of soft prompt parameters that learn to encode underlying information in the data relevant for unlearning. When prepended to the input tokens of an LLM during inference, the soft prompts guide the LLM towards a \textit{generic response}. We implement a multi-objective loss aligned with specific unlearning goals to facilitate the learning of soft prompts. SPUL unlearns undesirable outcomes without updating large-scale LLM parameters and can fully capitalize on the language understanding capability offered by the pre-trained LLMs. Consequently, our framework can utilize the same pre-trained LLM for different unlearning tasks and datasets during inference. 

We evaluate SPUL for sentiment classification and MCQA tasks on benchmark NLP datasets when unlearning a subset from the corresponding training dataset and compare against various fine-tuning-based methods. We show that SPUL can effectively induce forgetting during inference while preserving the pre-trained utility with significant improvements over baselines. We conduct experiments to analyze the influence of SPUL hyperparameters including the contribution of loss components and the size of the soft prompts. We further validate SPUL on multiple pre-trained LLMs of different parameter sizes and different sizes of unlearning sets.

\section{Related Work}
\subsection{Soft prompting}
Soft prompting or prompt tuning emerged as a lightweight alternative to fine-tuning while keeping pre-trained LLM parameters frozen. Motivated by discrete prompts that guide pre-trained LLMs via task-specific instructions or demonstration examples, soft prompting makes prompt design more efficient by employing trainable prompt parameters. The idea was conceived by~\citet{lester2021prompttuning}; they added trainable continuous embeddings to the encoder input sequence of an LLM and showed that the learned prompts achieve performance comparable to fine-tuning on NLP classification tasks with models having over 10B parameters. Simultaneously,~\citet{li2021ptuning} developed the notion of prefix tuning which prepends task-specific prefixes to the input embeddings along with the encoder and decoder inputs of an autoregressive LM and showed that their method is comparable to fine-tuning approaches for text generation tasks.~\citet{liu2021gptunderstands} concatenated trainable continuous prompts with discrete prompts along with a prompt encoder module that maps prompts to model inputs to improve performance on supervised and few-shot tasks. Subsequent research showed that deep prompt tuning achieves comparable performance to fine-tuning across several tasks on models of varying scales by inserting tunable parameters into every LLM layer~\citep{liu2022ptuning}.

\subsection{Unlearning in LLMs}
Machine unlearning arose as a promising solution to address data protection guidelines by efficiently forgetting training samples corresponding to unlearning requests in place of costly retraining~\citep{bourtoule2021unlearning, cao2015unlearning, liu2022federated, guo2020certified, sekhari2021unleanring, golatkar2020selective}. 
In the context of LLMs, machine unlearning is quickly gaining prominence due to concerns stemming from bias, toxicity, and privacy~\citep{si2023unlearning, liu2024rethinking}. Some works in this direction emphasize model parameter optimization via gradient ascent~\citep{jang2023unlearning, chen2023efficientunlearning, yao2024large, maini2024tofu, yao2024machine} to unlearn unwanted responses for specific examples or datasets. They also fine-tune the model with various knowledge alignment objectives to maintain model utility. Other works leverage parameter optimization via relabeling of unlearning data. For instance, ~\citet{eldan2023harry} unlearn Harry Potter content by fine-tuning the model via gradient descent to replace the model's response for queries related to Harry Potter with outputs containing generic translations. In contrast to these works,~\citet{jia2024soul} utilize similar fine-tuning objectives but focus on optimizer selection and propose a framework that performs influence-based model updates via second-order optimization. Additionally, some works propose localization-based objectives that aim to identify a subset of model units that represent information about unlearning data and effectively delete them~\citep{meng2022locating, yu2023partitioning, wu2023detecting}.
A few works also focus on modifying LLM input sequences to promote unlearning for black-box LLMs but are limited in the size of data that can be unlearned. For instance,~\citet{pawelczyk2023incontext} perform in-context unlearning by crafting input comprised of unlearn samples paired with flipped labels and other demonstrations with correct labels. ~\citet{thaker2024guardrail} investigate guardrail techniques for unlearning by instructing models to withhold unwanted knowledge or filtering undesirable LLM outputs.
Unlike most fine-tuning-based approaches, our goal in this work is to develop a soft prompting strategy to facilitate unlearning in LLMs. We aim to modulate LLM behavior using prompts similar to input modification strategies. However, instead of specifying manual instructions or providing demonstration samples as context, we leverage soft prompting to automate prompt optimization while adhering to unlearning objectives through loss specifications.

\section{Soft Prompting for Unlearning}

\subsection{Soft Prompting}\label{sec:soft_prompt}
Let $D = \{s_i, y_i\}_{i=1}^N$ denote a dataset containing $N$ input-output pairs where $s_i$ is a text sequence containing $n_i$ tokens and $y_i$ is the corresponding output. Also, let $h_\theta$ represent a pre-trained LLM with parameters $\theta$; $h_\theta$ can be prompted with $s_i$ to obtain an output $\hat{y}_i$. Assume $\mathbf{x}_i \in \mathbb{R}^{n_i\times d}$ denotes the token embeddings obtained for an arbitrary text sample $s_i$ from the embedding module of $h_\theta$ where $d$ is the dimension of the embedding space. We first define $p$ prompt tokens as $\boldsymbol{\boldsymbol{\phi}} = \{\phi_1,\cdots,\phi_p\}$ where $\phi_i \in \mathbb{R}^d$. To adapt $h_\theta$ over $D$ using soft prompts, $\boldsymbol{\phi}$ is appended to $\mathbf{x}_i$ to form the sequence $\{\boldsymbol{\phi}; \mathbf{x}_i\} \in \mathbb{R}^{(p+n_i)\times d}$ as input to the encoder or decoder in $h_\theta$. During backpropagation, the pre-trained parameters $\theta$ are frozen and gradient updates are applied only to $\boldsymbol{\phi}$ when maximizing the likelihood of the output $y_i$ as 
\begin{align}
    \argmax\limits_{\boldsymbol{\phi}}\log h_\theta(\{\boldsymbol{\phi}, \mathbf{x}_i\}).
\end{align}
The size of the learnable prompts $\boldsymbol{\phi}$ is very small compared to that of the pre-trained parameters $\theta$. Nonetheless, soft prompting has shown considerable performance over various language tasks with results comparable to fine-tuning. This motivates us to consider \textit{whether we can achieve unlearning in LLMs by optimizing continuous prompt tokens.}

\subsection{Problem Formulation}\label{sec:problem_def}
Given a training dataset $D^{tr}$ that was observed during pre-training of $h_\theta$, we assume a forget set, $D^{tr}_{f} \subset D^{tr}$, as the data intended for forgetting/removal from $h_\theta$. Simultaneously, we define a retain set $D^{tr}_{r} = D^{tr}\setminus D^{tr}_f$ comprising the remaining samples. Then, the goal of unlearning is to forget the token sequences in $D^{tr}_{f}$ while maintaining inference utility on $D^{tr}_r$. 
For our work, we focus on the task of text classification and question answering and interpret unlearning as the forgetting of the predictive output token sequences $y_i \in D^{tr}_f$. Essentially, we de-correlate text features and their corresponding labels for the relevant forget samples but preserve the predictive performance on the retain samples. To this end, we aim to design a soft prompting framework to obtain optimized prompt tokens that can guide the base model toward the forget and retain objectives. With our framework, we aim to address the following research questions. 

\noindent\textbf{RQ1:} How can soft prompting be utilized to effectively unlearn subsets of training data in the text classification/QA domain? \\
\noindent\textbf{RQ2:} How can soft prompting be implemented to achieve utility preservation with forgetting? \\
\noindent\textbf{RQ3:} How efficient is soft prompting-based unlearning compared to fine-tuning and re-training?

\subsection{Method}\label{sec:spul}
As soft prompts can be trained to encode signals from a dataset with the purpose of adapting a pre-trained LLM to a specific downstream task, we anticipate that the strategy can also be utilized to encode relevant information from an unlearning dataset containing forget and retain samples. Here, we propose the framework SPUL that leverages soft prompting to obtain effective prompt tokens $\boldsymbol{\phi}$ from an unlearning dataset $D^{tr}$. Since one of the unlearning objectives in our framework is to promote feature and text de-correlation for forget samples, we design a loss attuned to enforcing incorrect predictions for the respective text inputs. Specifically, we force the model to associate each input forget text sequence with a generic output token instead of its true label. We construct a generic label set $\bar{Y}$ that is disjoint from the task labels and contains tokens such as \textit{neutral}, \textit{unknown}, or \textit{None} and define a loss over the forget inputs,
\begin{align}
    \mathcal{L}_f &= \sum\limits_{(\mathbf{x}_i, .)\in D^{tr}_f} l(\hat{y_i}|\{\boldsymbol{\phi}, \mathbf{x}_i\}, \bar{y}_i),
\end{align}
where $\bar{y}_i$ denotes a uniform random sample drawn from the pre-defined generic label set $\bar{Y}$, and $l(\cdot)$ refers to the standard cross-entropy loss. Ideally, $\mathcal{L}_f$ allows the prompt tokens $\boldsymbol{\phi}$ to capture specific nuances from the samples in $D^{tr}_f$ and consequently guide the LLM to change its predictive sequence for an arbitrary example containing the learned distinctions. Simultaneously, unlearning also aims to preserve the predictive performance for samples not included in the forget set. In our SPUL framework, the prepended prompt tokens $\boldsymbol{\phi}$ should not change the predictive sequences for $\mathbf{x}_j \in D^{tr}_r$. Therefore, to preserve inference utility on the retain set, we define a loss using their true labels as
\begin{align}
    \mathcal{L}_r = \sum\limits_{(\mathbf{x}_j, y_j) \in D^{tr}_r}l(\hat{y}_j|\{\boldsymbol{\phi}, \mathbf{x}_j\}, y_j),
\end{align}
where $l(\cdot)$ again represents the cross-entropy loss. $L_r$ ensures that the model's utility on the retain set does not degrade with the addition of prompt tokens. In addition to maintaining performance on the retain set, the model after unlearning should closely resemble the model before unlearning. In our framework, we constrain the predictive distribution of the model such that $h_\theta(\{\boldsymbol{\phi}, \mathbf{x}_j\})$ reflects $h_\theta(\mathbf{x}_j)$ for any $\mathbf{x}_j \in D^{tr}_r$. We quantify this difference using KL divergence as
\begin{align}
    \mathcal{L}_{kl} = \sum\limits_{(\mathbf{x}_j, .) \in D^{tr}_r} \operatorname{KL}(h_\theta(\{\boldsymbol{\phi}, \mathbf{x}_j\}) || h_\theta(\mathbf{x}_j)),
\end{align}
where $\operatorname{KL}(\cdot)$ denotes the KL divergence term. $h_\theta(\{\boldsymbol{\phi}, \mathbf{x}_j\})$ represents the base model's predictive distribution conditioned on inputs prepended with the learnable prompt tokens and $h_\theta(\mathbf{x}_j)$ refers to the output distribution conditioned only on the input text sequence. We utilize $\mathcal{L}_{kl}$ in addition to $\mathcal{L}_r$ to avoid large deviations in the base model's output due to the influence from $L_f$. Finally, at each time step $t$ during training, we update $\boldsymbol{\phi}$ by optimizing the overall loss obtained as
\begin{align}
    \mathcal{L} = \mathcal{L}_f + \alpha\cdot\mathcal{L}_r + \beta\cdot\mathcal{L}_{kl},
\end{align}
where $\alpha$ and $\beta$ are hyperparameters that specify the contribution of the respective loss components. 

\begin{table}[t!]
    \centering
    \small
    \caption{Dataset Statistics}
    \begin{tabular}{ccccc}
    \toprule
        Dataset & $D^{tr}_f$ & $D^{tr}_r$ & $D^{te}_f$ & $D^{te}_r$ \\
    \midrule
        SST-2 & 1425 & 46331 & 610 & 19855 \\
        Yelp polarity & 5081 &  95012 & 885 & 18089 \\
        WMDP+SciQ & 900 & 12679 & 373 & 1000 \\ 
    \bottomrule
    \end{tabular}
    \label{tab:datasets}
\end{table}
\section{Experiments}
\subsection{Experimental Setup}
\paragraph{Datasets}\label{sec:dataset}
We evaluate SPUL on standard NLP datasets SST-2~\citep{socher2013sst} and Yelp polarity~\citep{zhang2015polarity} for the task of sentiment classification and datasets WMDP~\cite{li2024wmdp} and SciQ~\cite{welbl2017sciq} for the task of multiple-choice question answering. SST-2 and Yelp contain reviews with each text sequence being labeled as a positive or negative sentiment. SciQ consists of science exam questions about Physics, Chemistry, and Biology, among others in a four-way multiple-choice format where each answer choice is associated with symbols such as ``A'', ``B'', etc. and WMDP contains questions about hazardous knowledge in biosecurity, cybersecurity, and chemical security in the same format. To build a realistic unlearning scenario where unlearning requests from each user would likely include multiple related training samples, we preprocess the classification datasets to construct the forget and retain sets such that the forget samples are semantically similar to each other (Yelp) or refer to common entities (SST-2). For MCQA, we construct forget sets to unlearn potentially harmful information while retaining general science knowledge. 

For SST-2, we first perform Named Entity Recognition to identify named personalities, select a specific set of entities, and sample all related reviews to form the forget set $D^{tr}_f$. The remaining reviews are consequently assigned to the retain set $D^{tr}_r$. We perform a similar partitioning using the selected entities on the test set to obtain $D_f^{te}$ and $D_r^{te}$. 
For Yelp, we perform k-means clustering with cosine distance on the training data to divide the reviews into semantically similar groups. We randomly select a subset of the clusters and group them to form the $D^{tr}_f$ and the rest as $D^{tr}_r$. We utilize the same cluster centers to infer cluster identities for the test data and form the sets $D_f^{te}$ and $D_r^{te}$ accordingly. 
For the MCQA task, we construct the forget sets from WMDP containing questions about hazardous knowledge in biosecurity and the retain sets from SciQ with general science questions. We refer to this dataset as WMDP+SciQ.
The sizes of the constructed forget and retain sets are reported in~\cref{tab:datasets}.   

\begin{table*}[t!]
    \small
    \centering
    \caption{SPUL Unlearning performance compared to baselines}
    \setlength{\tabcolsep}{5pt}
    \begin{tabular}{ccccccccccccc}
    \toprule
    \multirow{2}{*}{Dataset} & \multirow{2}{*}{Method} & \multicolumn{2}{c}{Train Retain ($D^{tr}_r$)} & & \multicolumn{2}{c}{Train Forget ($D^{tr}_f$)} & & \multicolumn{2}{c}{Test Retain ($D^{te}_r$)} & & \multicolumn{2}{c}{Test Forget ($D^{te}_f$)}\\ 
    \cmidrule{3-4}\cmidrule{6-7}\cmidrule{9-10}\cmidrule{12-13}
     & & ACC(\%)$\uparrow$ & F1(\%)$\uparrow$ && ACC(\%)$\downarrow$ & F1(\%)$\downarrow$ && ACC(\%)$\uparrow$ & F1(\%)$\uparrow$ && ACC(\%)$\downarrow$ & F1(\%)$\downarrow$ \\
    \midrule
    \multirow{7}{*}{SST-2} & \cellcolor{red!25} Vanilla & 37.50 & 44.66 && 31.79 & 38.34 && 37.51 & 44.67 && 29.67 & 36.85 \\
    & \cellcolor{red!25} QLoRA & 99.89 & 99.89 && 99.72 & 99.72 && 95.57 & 95.57 && 96.07 & 96.07 \\ 
    & \cellcolor{yellow!25} GA & 55.66 & 39.80 && 53.93 & 37.83 && 55.96 &  40.16 && 56.89 & 41.25 \\
    & \cellcolor{yellow!25} RL & 33.31 & 48.08 && 13.82 & 22.97 && 31.00 & 45.56 && 14.26 & 24.18 \\
    & \cellcolor{yellow!25} GA+KL & 55.64 & 39.87 && 53.96 & 38.07 && 55.94 & 40.24 && 56.89 & 41.47 \\
    & \cellcolor{yellow!25} GA+GD & 97.17 & 97.50 && 13.75 & 20.58 && 94.43 & 94.76 && 11.31 & 17.18 \\
    & \cellcolor{green!15} SPUL & 99.15 & 99.39 && 12.98 &	22.94 && 94.93 & 95.24 && 16.07 & 27.42 \\ 
    \midrule
    \multirow{7}{*}{Yelp} & \cellcolor{red!25} Vanilla & 89.55 & 89.88 && 89.29 & 89.62 && 90.03 & 90.33 && 86.89 & 87.37 \\
    & \cellcolor{red!25} QLoRA & 99.31 & 99.31 && 99.49 & 99.49 && 98.42 & 98.41 && 98.76 & 98.76 \\
    & \cellcolor{yellow!25} GA & 66.11 & 63.48 && 67.90 & 64.62 && 65.13 & 62.37 && 67.91 & 64.24 \\
    & \cellcolor{yellow!25} RL & 53.00 & 67.75 && 52.84 & 66.78 && 52.75 & 67.40 && 49.94 & 65.01 \\
    & \cellcolor{yellow!25} GA+KL & 46.85 & 32.90 && 50.32 & 35.57 && 46.27 & 32.26 && 51.19 & 35.97 \\
    & \cellcolor{yellow!25} GA+GD & 99.23 & 99.42 && 79.69 & 86.98 && 97.76 & 98.00 && 80.90 & 88.19 \\
    & \cellcolor{green!15} SPUL & 89.74 & 93.43 && 55.03 & 70.48 && 89.63 & 93.29 && 60.23 & 74.69  \\ 
    \midrule
    \multirow{7}{*}{WMDP+SciQ} & \cellcolor{red!25} Vanilla & 47.75 & 46.85 && 26.78 & 17.46 && 46.20 & 46.11 && 23.59 & 14.86 \\
    & \cellcolor{red!25} QLoRA     & 99.74 & 99.74 && 98.11 & 98.11 && 91.80 & 91.80 && 62.73 & 62.83 \\
    & \cellcolor{yellow!25} GA	   & 99.35 & 99.35 && 86.89 & 87.44 && 90.70 & 90.71 && 57.64 & 58.66 \\
    & \cellcolor{yellow!25} RL	   & 99.32 & 99.32 && 84.11 & 89.57 && 90.40 & 90.40 && 53.35 & 59.54 \\
    & \cellcolor{yellow!25} GA+KL  & 98.84 & 98.85 && 67.44 & 68.91 && 90.20 & 90.24 && 49.33 & 50.26 \\
    & \cellcolor{yellow!25} GA+GD  & 99.42 & 99.42 && 27.22 & 13.38 && 90.00 & 90.02 && 22.25 & 8.84 \\
    & \cellcolor{green!15} SPUL    & 99.38 & 99.45 && 5.44 & 10.20 && 89.70 & 89.75 && 3.22 & 6.07 \\
    \bottomrule
    \end{tabular}
    \label{tab:main}
\end{table*}

\paragraph{Baselines}
We assess the effectiveness of SPUL by comparing its performance against multiple SOTA parameter-tuning baselines. Gradient Ascent (GA)~\citep{jang2023unlearning} optimizes pre-trained LLM parameters on the forget set by maximizing the cross-entropy loss in place of the standard minimization. Fine-tuning with Random Labels (RL)~\citep{golatkar2020selective, yao2024machine} similarly optimizes the base model on the forget set but by enforcing convergence on random labels. We use the generic label set discussed in~\cref{sec:spul} as the random labels for RL. Gradient Ascent + KL Divergence (GA + KL) and Gradient Ascent + Descent (GA+GD) integrate parameter optimization using the retain set with GA to balance forgetting effectiveness with utility~\cite{yao2024machine}. The former defines a KL-divergence constraint on the LLM's output distribution and the latter implements the standard cross-entropy loss. Note that for all four baselines, we perform full fine-tuning of the LLM following prior works based on their publicly available implementations. 

\paragraph{Settings}
We use LLaMA-2-7B~\cite{touvron2023LLaMA} as the base LLM to evaluate our SPUL framework. We further validate the unlearning effectiveness of our method with OPT-1.3B~\cite{zhang2022opt} and LLaMA-2-13B~\cite{touvron2023LLaMA}. To ensure familiarization with the unlearning dataset, we fine-tune the base LLMs on the full training dataset $D^{tr} = D^{tr}_f \cup D^{tr}_r$ for 10 epochs on SST-2, 2 epochs on Yelp, and 5 epochs on WMDP+SciQ with a learning rate set to 0.0001 and context length to 1024 using QLoRA~\citep{dettmers2023qlora}. We treat this fine-tuned version of the LLM as the base model for unlearning. As for the configurations of SPUL, we fix the learning rate at 0.0001 across all LLMs, and datasets and vary prompt token length $p$ among \{10, 20, 30, 40, 50\}. We also vary the regularization parameters $\alpha$ as \{0.1, 0.5, 1.0\} and $\beta$ as \{0.0, 0.1, 0.5, 1.0\}\footnote{We note that advanced approaches, e.g., utility function, Pareto-based, and constraint-based methods, can be potentially adopted to determine values of $\alpha$ and $\beta$.}. We train our unlearning framework for a total of 10 epochs. 
As for baseline model specifications, we follow earlier works~\cite{yao2024machine} and conduct a parameter search for the best learning rates to report the most competitive results after 1 epoch of training.  
All experiments are conducted on NVIDIA A100 GPUs with 40GB RAM and we report the evaluation metrics over a \textbf{single run} due to the resource-intensive nature of the experiments.

\begin{figure*}[h!]
    \centering
    \includegraphics[width=0.7\textwidth]{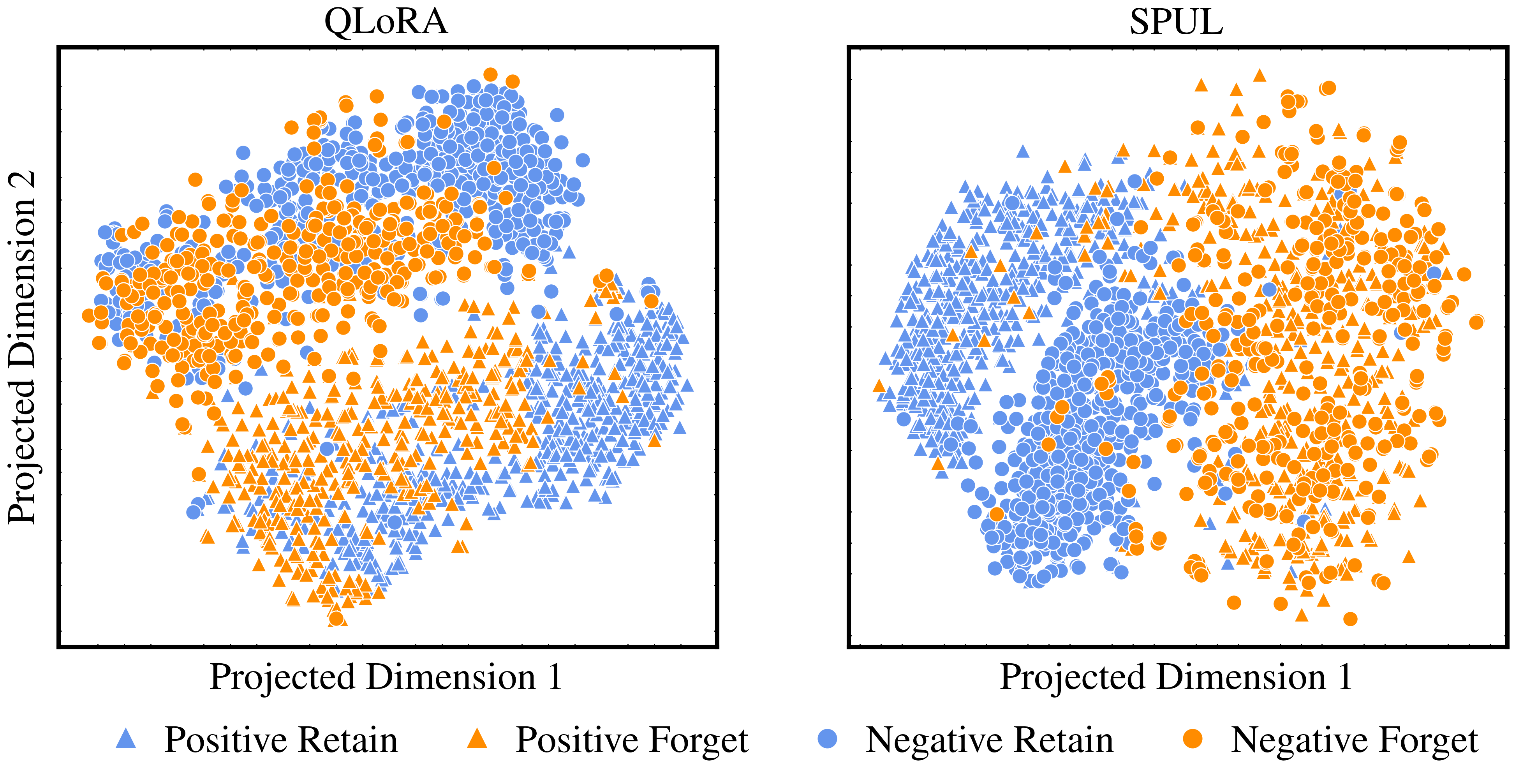}
    \caption{Embedding visualization results on SST-2 with QLoRA and SPUL}
    \label{fig:vis}
\end{figure*}

\paragraph{Evaluation}
We demonstrate the efficacy of the unlearning framework by evaluating the methods based on the research questions posed in~\cref{sec:problem_def}. To quantify how well our SPUL framework addresses RQ1, we report the accuracy and weighted F1 on the forget set, $D^{tr}_f$, which signifies whether the learned soft prompts can de-correlate the text features and labels. As $D^{te}_f$ is composed of text sequences semantically or lexically similar to $D^{tr}_f$, the prompt tokens should result in a comparable performance decline on $D^{te}_f$. To evaluate SPUL based on RQ2, we report model performance on $D^{tr}_r$ and consequently $D_r^{te}$. We emphasize the differences in the accuracy and F1 scores of the base model before and after unlearning to signify utility preservation. Finally, to answer RQ3, we report the number of training parameters and required GPU hours and compare them against baseline metrics.

\subsection{Experimental Results}

\paragraph{Main Results}
We include our main results with LLaMA-2-7B in~\cref{tab:main}. We report performance metrics for the original pre-trained LLM denoted as Vanilla and the fine-tuned base model denoted as QLoRA. We notice that the Vanilla results are considerably poorer for SST-2 compared to Yelp which validates our setup of fine-tuning the original LLM on the datasets for memorization. We attribute the difference in utility to the fact that the text sequences in Yelp are significantly longer and provide more contextual information. Nonetheless, after fine-tuning with QLoRA, the LLM's performance increases to similar margins for both datasets. QLoRA fine-tuning similarly improves LLM's predictions for the WMDP+SciQ dataset. 

From~\cref{tab:main}, we observe that SPUL significantly reduces accuracy and F1 on $D^{tr}_f$ compared to QLoRA demonstrating forgetting efficiency. At the same time, the difference in utility between SPUL and QLoRA for $D^{tr}_r$ is minimal showing that our method can promote unlearning while also preserving inference utility. Moreover, the metrics for $D^{te}_f$ and $D^{te}_r$ reflect those reported for $D^{tr}_f$ and $D^{tr}_r$ showing that the soft prompts effectively impose unlearning constraints on samples unseen during training. We observe similar performance trends for Yelp and WMDP+SciQ. Although the performance drop for $D_f^{tr}$ and $D_r^{te}$ in Yelp are not equally as large as SST-2, the forget utility with the learned tokens is significantly lesser in comparison to the base model. We conjecture that the additional context provided by descriptive Yelp reviews restricts the forgetting capacity of the LLM. Also point out that utility loss in retain sets is smaller than forget sets. We also notice that SPUL performs exceedingly well on the WMDP+SciQ with the highest differences between the retain and forget metrics.

Furthermore, SPUL outperforms baseline methods by a significant margin; compared to GA and RL, which optimize model parameters based only on the $D_f^{tr}$, SPUL consistently preserves inference utility on the retain sets with comparable or even lower metrics on the forget set. For WMDP+SciQ, both baselines underperform in the forgetting task. GA+KL and GA+GD optimize model parameters based on both $D_f^{tr}$ and $D_r^{tr}$. However, GA+KL performs poorly on all three datasets. GA+GD performs especially well on SST-2 and WMDP+SciQ but fails to enhance forget quality on Yelp which has more descriptive reviews than SST-2. The proposed SPUL framework can however attain effective unlearning with the least loss of model utility. Among the compared methods, SPUL achieves significantly better overall trade-offs between the contrasting unlearning objectives of performance degradation and utility preservation for both tasks. 
\begin{table*}[t!]
    \small
    \centering
    \caption{SPUL performance on SST-2 across varying $\alpha$ and $\beta$ values at $p = 30$}
    \setlength{\tabcolsep}{5pt}
    \begin{tabular}{ccccccccccccc}
    \toprule
    \multirow{2}{*}{$\alpha$} & \multirow{2}{*}{$\beta$} & \multicolumn{2}{c}{Train Retain ($D^{tr}_r$)} & & \multicolumn{2}{c}{Train Forget ($D^{tr}_f$)} & & \multicolumn{2}{c}{Test Retain ($D^{te}_r$)} & & \multicolumn{2}{c}{Test Forget ($D^{te}_f$)}\\
    \cmidrule{3-4}\cmidrule{6-7}\cmidrule{9-10}\cmidrule{12-13}
     & & ACC(\%)$\uparrow$ & F1(\%)$\uparrow$ && ACC(\%)$\downarrow$ & F1(\%)$\downarrow$ && ACC(\%)$\uparrow$ & F1(\%)$\uparrow$ && ACC(\%)$\downarrow$ & F1(\%)$\downarrow$ \\
    \midrule
    \multirow{4}{*}{0.1} & 0.0 & 90.84 & 92.69 && 9.12  & 16.55 &&	89.50 &	91.15 &&	10.33 &	18.40 \\
                         & 0.1 & 92.59 & 93.75 && 6.81  & 12.62 &&	90.77 &	91.85 &&	10.16 &	18.29 \\
                         & 0.5 & 96.77 & 97.91 && 8.70  & 15.98 &&	93.01 &	94.10 &&	11.15 &	19.81 \\
                         & 1.0 & 85.19 & 88.00 && 8.49  & 15.47 &&	84.64 &	87.19 &&	10.66 &	19.02 \\
    \midrule
    \multirow{4}{*}{0.5} & 0.0 & 98.17 & 98.69 && 11.86 &	21.17 &&	94.34 &	94.87 &&	14.59 &	25.07 \\
                         & 0.1 & 97.57 & 97.95 && 11.09 &	19.88 &&	94.22 &	94.58 &&	11.97 &	21.08 \\
                         & 0.5 & 97.74 & 98.35 && 13.82 &	24.21 &&	93.97 &	94.57 &&	17.21 &	29.08 \\
                         & 1.0 & 93.87 & 94.66 && 11.51 &	20.39 &&	91.62 &	92.36 &&	14.59 &	25.03 \\
    \midrule
    \multirow{4}{*}{1.0} & 0.0 & 97.52 & 97.91 && 12.14 &	21.60 &&	94.22 &	94.65 &&	15.57 &	26.50 \\
                         & 0.1 & 98.64 & 98.96 && 12.14 &	21.54 &&	94.63 &	94.97 &&	16.07 &	27.41 \\
                         & 0.5 & 99.15 & 99.39 && 12.98 &	22.94 &&	94.93 &	95.24 &&	16.07 &	27.42 \\
                         & 1.0 & 95.70 & 96.19 && 14.88 &	25.75 &&	93.05 &	93.55 &&	17.38 &	29.18 \\
    \bottomrule
    \end{tabular}
    \label{tab:ablation_alpha_beta}
\end{table*}

\paragraph{Visualization}
We also visualize model outputs to show the effectiveness of our SPUL method. We utilize outputs from the last embedding layer of the LLM and map them onto a t-SNE diagram as shown in~\cref{fig:vis}. The plots represent 500 data points randomly sampled from the training dataset in SST-2 for each label. In the plots, we use colors to differentiate the retain and forget examples and use shapes to differentiate the positive and negative examples. We visualize the embeddings from QLoRA, i.e., the base model before unlearning and we observe a clear divide between the positively and negatively labeled samples in the embedding space. The retain and forget samples are clustered together within the regions defined by each label. For the t-SNE plot of SPUL, i.e., the embeddings obtained after pretending the learned soft prompts, we notice a clear separation between the retain and forget samples as indicated by the blue and orange regions in~\cref{fig:vis}. This shows that the soft prompts truly capture the differences between the forget and retain sets. Moreover, the retain samples are further grouped into clusters per their labels. On the other hand, the positive and negative forget samples are mixed together. This shows that the soft prompt tokens learned by SPUL successfully guide the LLM to unlearn text and label correlation for the forget samples while preserving predictive utility on the retain set. 

Referring back to~\cref{tab:main}, SPUL metrics on $D^{tr}_f$ and $D^{te}_f$ closely resemble each other for both SST-2 and Yelp. We make similar observations for $D^{tr}_r$ and $D^{te}_r$. Our visualization results also show that the output embeddings for forget samples are not distinguishable between labels. Compared to QLoRA visualization, model outputs for positive and negative retain samples are closer in the embedding space as well. As a result, in a black-box Membership Inference Attack (MIA)~\citep{shokri2017mia} scenario, it would be challenging to infer whether a particular forget sample was observed during training based only on model outputs.
\begin{table*}[!t]
    \small
    \centering
    \caption{SPUL performance on SST-2 across varying sizes of forget sets}
    \begin{tabular}{cccccccccccc}
    \toprule
    \multirow{2}{*}{$\tau$} & \multicolumn{2}{c}{Train Retain ($D^{tr}_r$)} && \multicolumn{2}{c}{Train Forget ($D^{tr}_f$)} && \multicolumn{2}{c}{Test Retain ($D^{te}_r$)} && \multicolumn{2}{c}{Test Forget ($D^{te}_f$)}\\
    \cmidrule{2-3}\cmidrule{5-6}\cmidrule{8-9}\cmidrule{11-12}
     & ACC(\%)$\uparrow$ & F1(\%)$\uparrow$ && ACC(\%)$\downarrow$ & F1(\%)$\downarrow$ && ACC(\%)$\uparrow$ & F1(\%)$\uparrow$ && ACC(\%)$\downarrow$ & F1(\%)$\downarrow$ \\
    \midrule
    25\%  & 99.37 & 99.60 && 26.69 & 42.07 && 95.10 & 95.38 && 39.84 & 56.22 \\
    50\%  & 97.66 & 98.47 && 18.96 & 31.78 && 93.80 & 94.62 && 23.61 & 37.60 \\
    100\% & 95.70 & 96.19 && 14.88 & 25.75 && 93.05 & 93.55 && 17.38 & 29.18 \\
    \bottomrule
    \end{tabular}
    \label{tab:forget_size}
\end{table*}
\begin{figure}[h!]
    \centering
    \includegraphics[width=0.49\textwidth]{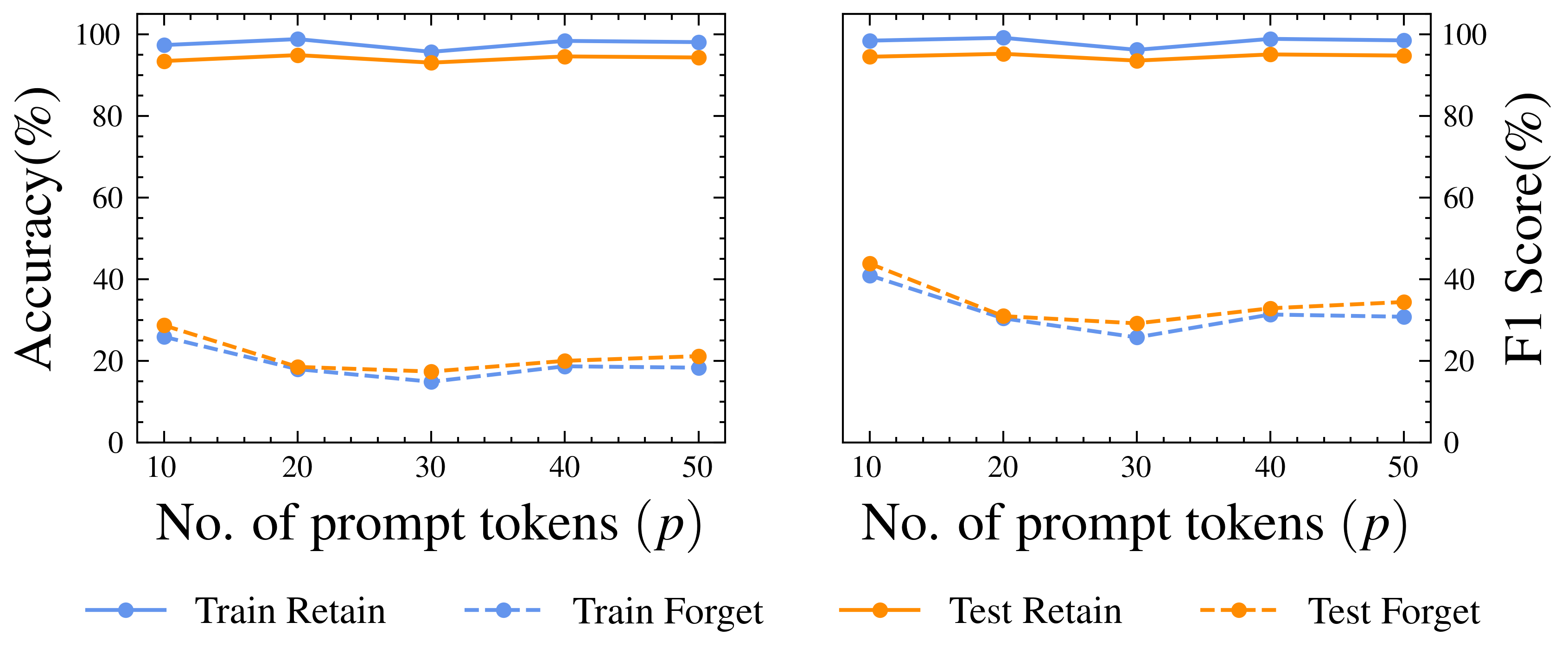}
    \caption{SPUL performance on SST-2 across varying $p$ at $\alpha = 1$ and $\beta = 1$}
    \label{fig:ablation_num_tokens}
\end{figure}

\paragraph{Ablation Study}
We conduct a series of experiments to investigate the influence of $\mathcal{L}_r$ and $\mathcal{L}_{kl}$ on the unlearning performance of the proposed SPUL framework and report the results in~\cref{tab:ablation_alpha_beta} for the SST-2 dataset. The hyperparameters $\alpha$ and $\beta$ control the influence of the retain set on the learned soft prompts via losses $\mathcal{L}_r$ and $\mathcal{L}_{kl}$ respectively. We fix the number of prompt tokens $p$ at 30 for all results and vary $\alpha$ in \{0.1, 0.5, 1.0\} and $\beta$ among \{0.0, 0.1, 0.5, 1.0\}. From~\cref{tab:ablation_alpha_beta}, we observe that at a fixed $\alpha$, unlearning efficacy is fairly unaffected by the change in the value of $\beta$. Model utility on the retain set, however, slightly increases as $\beta$ increases from 0.0 to 0.5 as $\mathcal{L}_{kl}$ gets more significance in the overall loss. We generally observe the best retain performance at $\beta$ = 0.5. The value of $\alpha$ influences performance on both forget and retain sets; higher $\alpha$ values benefit retain performance by prioritizing utility preservation whereas lower $\alpha$ values improve unlearning efficacy.      

\paragraph{Hyperparameter Study}
We also study the effect of the number of prompt tokens, represented by $p$, on the unlearning effectiveness of SPUL. We fix both $\alpha$ and $\beta$ at 1 and run experiments with $p$ ranging from 10 to 50 on SST-2 and report results in ~\cref{fig:ablation_num_tokens}. We find that inference utility on retain sets $D_r^{tr}$ and $D_r^{te}$ is largely unaffected by the different choice of $p$. However, we observe the most competitive forget performance at $p$ = 30 with increasing accuracy and F1 as $p$ increases/decreases. We speculate that the soft prompts mostly encode information from the forget set, for instance, the named entities in SST-2 whose reviews are unlearned, and ultimately instruct the LLM to misclassify examples with similar encodings. Accordingly, a larger $p$ generally benefits our soft prompting framework as made evident by the decline in forget metrics but may require longer training for optimal performance.  

\begin{table*}[!t]
    \small
    \centering
    \caption{SPUL performance on SST-2 dataset using OPT-1.3B and LLaMA-2-13B}
    \setlength{\tabcolsep}{4pt}
    \begin{tabular}{ccccccccccccc}
    \toprule
    \multirow{2}{*}{LLM} & \multirow{2}{*}{Method} & \multicolumn{2}{c}{Train Retain ($D^{tr}_r$)} && \multicolumn{2}{c}{Train Forget ($D^{tr}_f$)} && \multicolumn{2}{c}{Test Retain ($D^{te}_r$)} && \multicolumn{2}{c}{Test Forget ($D^{te}_f$)}\\
    \cmidrule{3-4}\cmidrule{6-7}\cmidrule{9-10}\cmidrule{12-13}
    & & ACC(\%)$\uparrow$ & F1(\%)$\uparrow$ && ACC(\%)$\downarrow$ & F1(\%)$\downarrow$ && ACC(\%)$\uparrow$ & F1(\%)$\uparrow$ && ACC(\%)$\downarrow$ & F1(\%)$\downarrow$ \\
    \midrule
    \multirow{3}{*}{OPT-1.3B} & \cellcolor{red!25}  Vanilla & 3.05 & 5.68 && 1.68 & 3.20 && 3.24 & 6.03 && 3.28 & 6.08 \\
    & \cellcolor{red!25} QLoRA   & 99.47 & 99.47 && 99.16 & 99.16 && 95.39 & 95.39 && 95.25 & 95.25 \\
    & \cellcolor{yellow!25} GA		& 79.50 & 78.01 && 70.67 & 67.02 && 78.70 & 77.05 && 71.97 & 67.99 \\
    & \cellcolor{yellow!25} RL		& 55.66 & 39.80 && 53.96 & 37.83 && 55.96 & 40.16 && 56.89 & 41.25 \\
    & \cellcolor{yellow!25} GA+KL	& 81.30 & 80.15 && 60.49 & 50.94 && 79.08 & 77.60 && 64.75 & 56.74 \\
    & \cellcolor{yellow!25} GA+GD	& 87.56 & 87.59 && 50.53 & 49.07 && 86.47 & 86.50 && 55.25 & 54.59 \\
    & \cellcolor{green!15} SPUL & 94.87 & 96.89 && 16.84 & 28.74 && 91.65 & 93.51 && 17.87 & 29.84 \\
    \midrule 
    \multirow{3}{*}{LLaMA-2-13B} & \cellcolor{red!25} Vanilla & 61.04 & 70.96 && 59.65 & 69.51 && 60.32 & 70.38 && 59.18 & 68.79 \\
    & \cellcolor{red!25} QLoRA   & 99.48 & 99.48 && 99.30 & 99.30 && 96.02 & 96.02 && 95.90 & 95.90 \\
    & \cellcolor{green!15} SPUL & 98.87 & 98.93 && 5.97 & 11.25 && 95.50 & 95.60 && 7.38 & 13.54 \\    
    \bottomrule
    \end{tabular}
    \label{tab:model}
\end{table*}

\paragraph{Forget Set Size}
To demonstrate the stability of our method w.r.t. the size of forget data, we evaluate SPUL on varying sizes of the train forget set $D_f^{tr}$ by sub-sampling $\tau$ = \{25\%, 50\%, 100\%\} of the original forget set constructed for SST-2. For the test forget set $D^{te}_f$ and the retain sets $D^{ts}_r$ and $D^{te}_r$, we use the same sets defined in~\cref{sec:dataset} for all three configurations of $D^{tr}_f$ to facilitate comparison. We present the results from this experiment on SST-2 in~\cref{tab:forget_size}. Our results indicate that SPUL can achieve utility preservation across differing numbers of forget samples with minimal loss as more forget samples are added to $D^{tr}_f$. In contrast to the retain metrics, SPUL clearly performs better for the forget metrics when more forget samples are present in the data for SST-2. Experimental results on Yelp presented in~\cref{tab:main} also highlight the robustness of SPUL against large forget sets as we assign more than 5000 samples to $D^{tr}_f$. As the training data contains comparatively fewer forget samples than retain samples, having a larger $D_f^{tr}$ allows the framework to emphasize the forgetting objective thus improving the unlearning efficacy.

\paragraph{Results on LLaMA-2-13B and OPT-1.3B}
We additionally evaluate the unlearning efficacy of our SPUL on different LLMs. In particular, we purposely choose OPT-1.3B with fewer parameters and LLaMA-2-13B with almost double the parameters compared to LLaMA-2-7B. In addition to the unlearning efficacy, this study also evaluates the scalability of our SPUL framework. We fix the hyperparameters $\alpha$ and $\beta$ at 1 and $p$ at 30 and report the results for SST-2 in~\cref{tab:model}. We first observe that the Vanilla inference with OPT-1.3B model performs noticeably poorer than LLaMA-2-7B whereas LLaMA-2-13B significantly improves over the initial metrics. This may be attributed to the pre-trained models' complexity which affects their generalization ability. We similarly perform fine-tuning using QLoRA to ensure the respective LLM has memorized the unlearning dataset. Moreover, SPUL can effectively achieve the forget and retain unlearning objectives as made evident by the low forget accuracy and F1 compared to the retain metrics that closely resemble the base model's performance. The results also indicate that the larger the LLM, the better it adapts to the unlearning task in our SPUL framework. Additionally, our results with OPT-1.3B indicate that SPUL significantly outperforms all baseline methods for both retain and forget metrics. We note that we could not run experiments on baselines with LLaMA-2-13B due to limited GPU as all baseline methods require full fine-tuning of the LLM. This further highlights the advantage of SPUL over baselines in terms of parameter efficiency.

\paragraph{Efficiency}
For LLMs, retraining from scratch is practically infeasible due to computational time and resources required for a huge set of parameters. Although fine-tuning pre-trained LLMs incurs less costs than retraining, the cost is still high. For instance, the LLM architectures used in our experiments require gradient updates for 1.42B, 6.74B, and 13B parameters for OPT-1.3B, LLaMA-2-7B, and LLaMA-2-13B respectively when implementing unlearning based on fine-tuning. When $p=30$, our SPUL reduces the computation cost by only optimizing 604K, 1.19M, and 1.49M parameters while freezing LLM parameters. Further increasing $p$ only linearly scales the number of training parameters. We also look at the running time of SPUL on the SST-2 compared against baseline methods and find the execution time required by each model of SPUL, GA + KL, and GA+GD for one training epoch is fairly similar, around 1020 GPU seconds, as SPUL also accesses LLM parameters during backpropagation. GA and RL methods are much quicker with approximate 40 GPU seconds of per epoch training time as these methods only consider the forget set. Nonetheless, SPUL avoids the overhead associated with updating LLM parameters, making it more resource-efficient.

\section{Conclusion}
In this work, we investigate unlearning in LLMs to remove the influence of unwanted training examples during text classification and multiple-choice question answering. We present a soft prompting strategy to unlearn subsets of training data while keeping pre-trained LLM parameters frozen to maintain the model's generalizability. Our SPUL framework optimizes a small number of prompt tokens using a multi-objective loss function defined on disjoint training data subsets representing the forget data that is subjected to removal and the retain data that aims to preserve model utility. Experimental evaluation on sentiment classification and QA datasets demonstrates the superior efficiency of our soft prompting-based unlearning over fine-tuning-based baselines. We also empirically show that SPUL can adapt to multiple LLMs and is robust to a high number of unlearning samples.

\section*{Acknowledgements}
This work was supported in part by NSF grants 1946391 and 2119691. 

\section*{Limitations}
We address the limitations of this work in the following. 
Our experiments primarily focus on open-source LLMs as the soft prompting framework requires access to frozen pre-trained parameters to compute gradients for the soft prompts despite not needing to update the LLM parameters.
Furthermore, this work focuses on the task of text classification, specifically sentiment classification, and question answering for the formulation of the unlearning framework and evaluation. Future research could explore the efficiency of soft prompting to achieve unlearning in the context of NLP tasks such as text generation, text summarization, and so on. Also, the soft prompting unlearning framework has not been evaluated comprehensively as we emphasize performance metrics to demonstrate unlearning efficacy. We note that there is a lack of an extensive evaluation pipeline for LLM unlearning in the current literature. Further research is needed to evaluate the robustness of the framework subject to model-stealing attacks, MIAs, and jailbreaking attempts.

\section*{Broader Impacts}
In this study, our focus is to achieve LLM unlearning in a resource-efficient manner. We aim to enable forgetting of unwanted and undesirable knowledge as per users' requests while maintaining model efficiency to avoid exploitation of sensitive information.
The datasets used for evaluation are publicly available and implemented within the intended use. Our usage of publicly available pre-trained LLMs also adheres to the associated licenses. We hope our study can further the research and literature on resource-efficient LLM unlearning.

\end{document}